\documentclass{article}
\usepackage{spconf,amsmath,epsfig}
\usepackage{appendix}
\usepackage{times}
\usepackage{soul}
\usepackage{url}
\usepackage[hidelinks]{hyperref}
\usepackage[utf8]{inputenc}
\usepackage[small]{caption}
\usepackage{graphicx}
\usepackage{amsmath}
\usepackage{amsthm}
\usepackage{booktabs}
\usepackage{makecell}
\usepackage{algorithm}
\usepackage{algorithmic}
\usepackage{tabu}
\usepackage{booktabs}
\usepackage{multirow}
\usepackage{subfigure}

\usepackage[table,xcdraw]{xcolor}
\usepackage{breqn}
\usepackage{blindtext}

\usepackage{amssymb}
\usepackage{bbm}

\copyrightnotice{\textbf{978-1-6654-3864-3/21/\$31.00~\copyright 2021 IEEE}}

\let\OLDthebibliography\thebibliography
\renewcommand\thebibliography[1]{
  \OLDthebibliography{#1}
  \setlength{\parskip}{0pt}
  \setlength{\itemsep}{0pt plus 0.3ex}
}
\usepackage{fancyhdr}

\begin{document}\sloppy

\def\x{{\mathbf x}}
\def\L{{\cal L}}

\title{Towards Intrinsic Common Discriminative Features Learning for Face Forgery Detection using Adversarial Learning}
%
\name{Wanyi Zhuang, Qi Chu$^{\ast}$, Haojie Yuan, Changtao Miao, Bin Liu and Nenghai Yu}
\address{CAS Key Laboratory of Electromagnetic Space Information, University of Science and Technology of China \\
wy970824@mail.ustc.edu.cn, qchu@ustc.edu.cn, \{doubihj, miaoct\}@mail.ustc.edu.cn,\\ \{flowice, ynh\}@ustc.edu.cn.}

\maketitle
\renewcommand{\thefootnote}{\fnsymbol{footnote}}
\footnotetext[1]{Corresponding authors.} 

%
\begin{abstract}
Existing face forgery detection methods usually treat face forgery detection as a binary classification problem and adopt deep convolution neural networks to learn discriminative features. The ideal discriminative features should be only related to the real/fake labels of facial images. However, we observe that the features learned by vanilla classification networks are correlated to unnecessary properties, such as forgery methods and facial identities. Such phenomenon would limit forgery detection performance especially for the generalization ability.
Motivated by this, we propose a novel method which utilizes adversarial learning to eliminate the negative effect of different forgery methods and facial identities, which helps classification network to learn intrinsic common discriminative features for face forgery detection. 
To leverage data lacking ground truth label of facial identities, we design a special identity discriminator based on similarity information derived from off-the-shelf face recognition model. 
Extensive experiments demonstrate the effectiveness of the proposed method under both intra-dataset and cross-dataset evaluation settings.
\end{abstract}
\section{Introduction}\label{sec:intro}
With the recent progress in synthetic image generation and manipulation, various advanced face forgery technologies have emerged, such as face swapping\cite{FaceSwap2019}, face reenactment\cite{thies2016face2face} and so on.
%
It leads to generating highly realistic fake human faces that can deceive human beings more easily. 
These forged facial images may be abused for malicious purposes, which inevitably brings serious security and privacy concerns, \textit{e.g.} fake news and evidence. Therefore, it's of great significance to develop powerful techniques to detect fake faces.

\begin{figure}
    \centering
    \begin{minipage}{0.48\linewidth}
    \centerline{\includegraphics[width=1\textwidth]{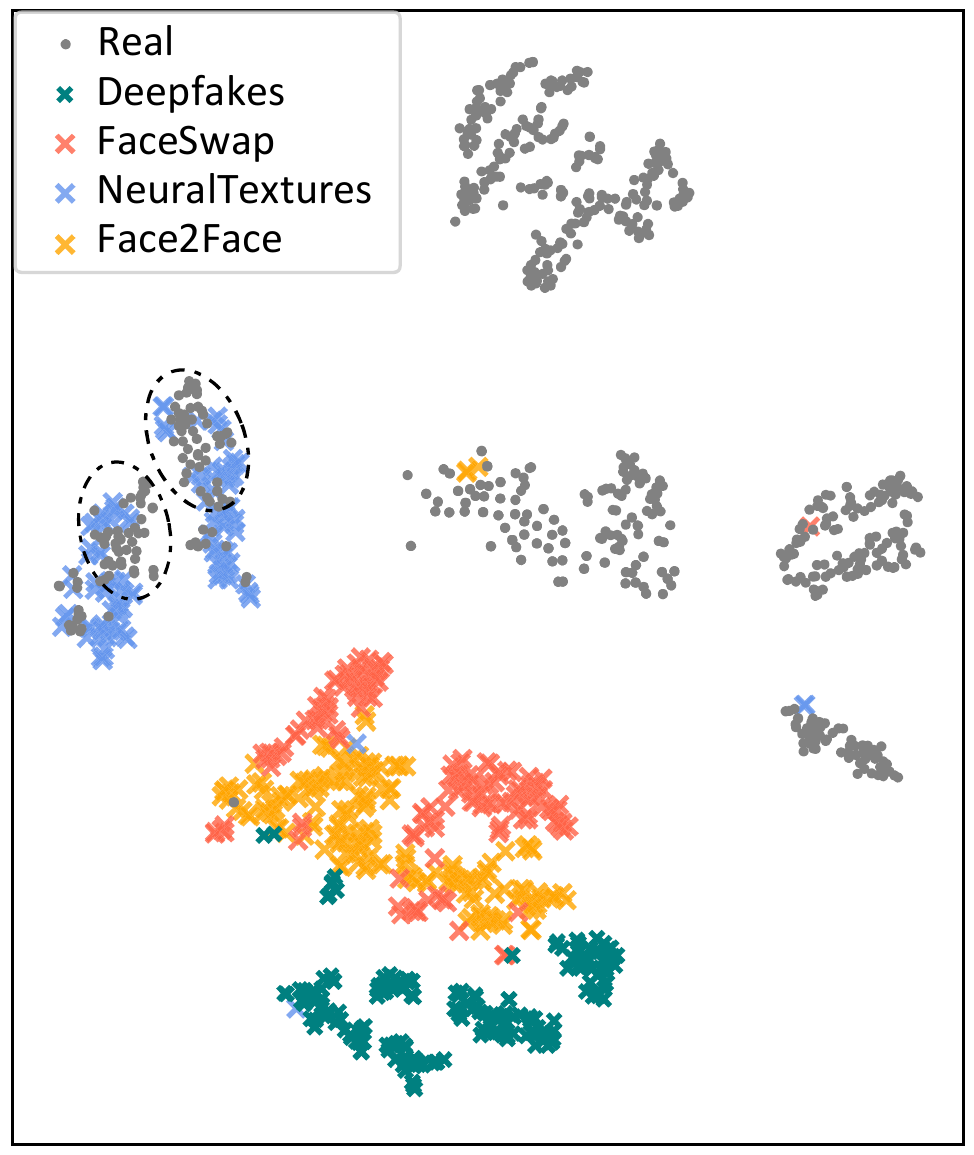}}
    \end{minipage}
    \begin{minipage}{0.48\linewidth}
    \centerline{\includegraphics[width=1\textwidth]{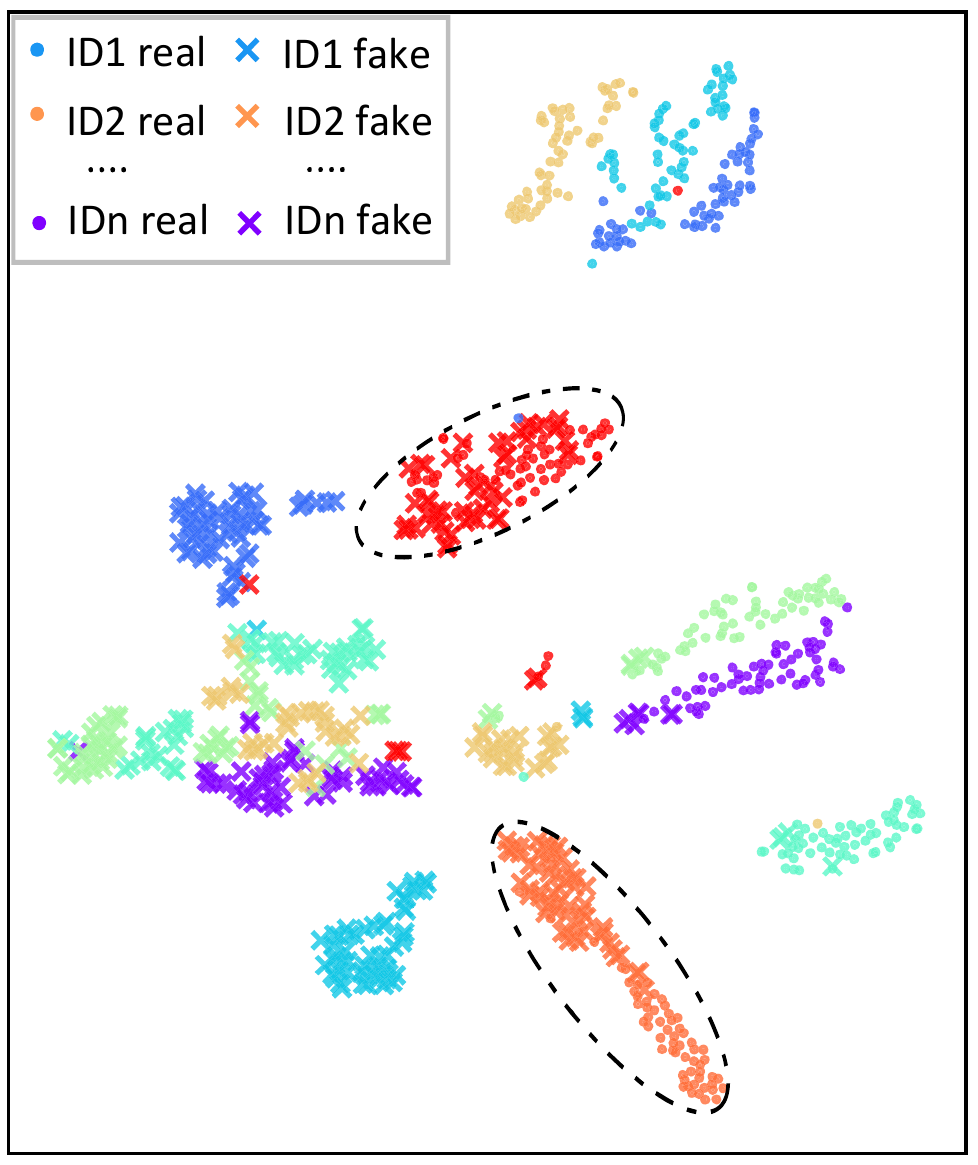}}
    \end{minipage}
    \caption{The t-SNE visualization of the features extracted by Xception\cite{chollet2017xception} trained on FaceForensics++\cite{rossler2019faceforensics++}. 
    The dots and forks stand for features of real and fake images respectively. The circled regions stand for misclassified samples. Left: The effect of face forgery methods.
    Right: The effect of facial identities. 
    Zoom in for details. }
    \vspace{-15pt} 
\label{motivation figure}
\end{figure}

To this end, many face forgery detection methods (\cite{chollet2017xception, nguyen2019capsule, dang2020detection,Zhao_2021_CVPR}) have been proposed. Existing methods\cite{chollet2017xception, nguyen2019capsule} usually treat face forgery detection as a binary classification problem and adopt deep neural networks to learn discriminative features. Despite their impressive performance, we observe that the features learned by these vanilla networks are related to some extra information that is irrelevant to real/fake classification but exists in the dataset.
Fig.\ref{motivation figure} illustrates the t-SNE visualization of the features extracted by Xception\cite{chollet2017xception} trained on FaceForensics++ dataset\cite{rossler2019faceforensics++}. Ideally, the intrinsic common discriminative features for face forgery detection should be only related to the real/fake labels of the facial images. However, as shown in Fig.\ref{motivation figure}, the features extracted by Xception can be clustered according to the face forgery methods (Left) and facial identities (Right) respectively. These phenomena indicate that the features learned by the vanilla classification network are not intrinsic enough for discriminating real/fake facial images, which would cause the features of real and fake images indistinguishable in some cases (the regions marked with circles in Fig.\ref{motivation figure}) and thus limits the final performance especially for generalization ability.

The above observations are consistent with the opinions of some recent works. DeepFakes Detection Challenge (DFDC) \cite{dolhansky2020deepfake} concluded that using a form of data augmentation that drops discriminative face parts (eyes, mouth, forehead, etc.) to mitigate overfitting is a common strategy of the top-performing models. It indicates that the vanilla classification network learns to capture the high-level semantic information related to facial identities which limits the generalization ability. A recent work\cite{luo2021generalizing} claimed that deep CNN models learn to capture the \textit{method-specific} texture patterns for forgery detection. Since there are many different manipulation algorithms, it is hard for the vanilla classification model to capture common discriminative features from different kinds of forgeries.
%

Based on our observations and prior works, we propose a novel face forgery detection method, which aims to learn the intrinsic common discriminative features via adversarial learning. Specifically, we introduce two discriminators for face forgery method classification and facial identity recognition respectively. In the case of lacking ground truth label of facial identities, we could not adopt the simple multi-class classification network as the discriminator for face identity recognition. To handle this problem, we design a special identity discriminator based on similarity information derived from off-the-shelf face recognition model. 
Extensive experiments show that
with the help of adversarial learning between the feature extractor and these two discriminators, the network learns to extract more intrinsic common discriminative features for face forgery detection by eliminating the negative effect of different forgery methods and facial identities.

\vspace{-5pt}
\section{Related Works}

\subsection{Face Forgery Detection Methods}
In the early years, face manipulation techniques usually produced face forgery images with obvious artifacts or inconsistencies, which could be leveraged as important cues for face forgery detection. For example,
Li \textit{et al.}\cite{li2018ictu} observed that the blinking frequency of the forgery video is lower than the normal, leading to the discriminative clues for forgery detection. 
Similarly, Matern \textit{et al.}\cite{matern2019exploiting} used hand-crafted visual features in eyes, noses, teeth to distinguish the fake faces.

Since there are universal procedures in different face manipulation methods, another series of works attempted to find artifacts in these common steps, such as affine warping (DSP-FWA\cite{li2018exposing}) and blending (Face X-ray\cite{li2020face}).  However, these methods suffer from great performance drops on low quality images, since high compression destroy these artifacts clues.

With the development of large-scale face forgery datasets \cite{rossler2019faceforensics++,li2020celeb,dang2020detection}, data-driven deep learning based methods emerged to adopt classification networks to directly learn the discriminative features for detecting fake faces, which currently dominate face forgery detection. 
Various deep neural networks were used to extract high level discriminative features. A line of works such as Capsule Network\cite{nguyen2019capsule} and XceptionNet\cite{rossler2019faceforensics++},  had achieved high detection accuracy on intra-dataset face forgery detection task.
However, these methods do not consider the effect of some extra information in the dataset such as the forgery methods and facial identities, which causes the learned features are related to these extra information as shown in Fig.\ref{motivation figure}.
Recently, some methods attempted to introduce frequency clues, such as Two-branch\cite{masi2020two}, F3-Net\cite{qian2020thinking}, FDFL\cite{Li_2021_CVPR} 
or attention mechanism (Xception+Reg.\cite{dang2020detection}, Multi-attention\cite{Zhao_2021_CVPR},\cite{miao2021TBIOM}) into face forgery detection to better detect manipulated faces. 
Different from these methods, we attempt to adopt adversarial learning to eliminate the negative effects of different forgery methods and facial identities for learning the intrinsic common discriminative features.

\vspace{-6pt}
\subsection{Adversarial Learning}
Adversarial learning, which generally contains two networks (a generator and a discriminator) with competing goals, has been used in various applications, such as image synthesis \cite{dong2017semantic}, domain generalization\cite{li2018domain} and so on. Recently, Jia et al.\cite{jia2020single} proposed a single-side domain generalization framework for face anti-spoofing using adversarial learning. Although we both adopt adversarial learning for learning better features, our method is quite different from \cite{jia2020single}. Firstly, our method aims to learn intrinsic common discriminative features that are irrelevant to face forgery methods. While they attempt to learn a feature space where the distribution of the real faces is compact while that of the fake ones is dispersed among domains. Besides, we design a novel identity discriminator to take facial identities into consideration, which could handle the case of lacking ground truth label of facial identities. While the identity information is not considered in \cite{jia2020single}.

\section{Methodology}

\subsection{Overview}

Based on vanilla binary classifier, we introduce two adversarial learning components:1) forgery method adversarial learning that attempts to make different forgery methods indistinguishable; 2) identity adversarial learning that attempts to make facial identities indistinguishable. 
Specifically, a forgery method discriminator and a identity discriminator are embedded after the feature generator as shown in Fig.~\ref{fig:framework}.


\vspace{-6pt}
\subsection{Forgery Method Adversarial Learning}\label{forgery illustration}
\begin{figure}[!t]
    \centering
    \includegraphics[width=0.50\textwidth]{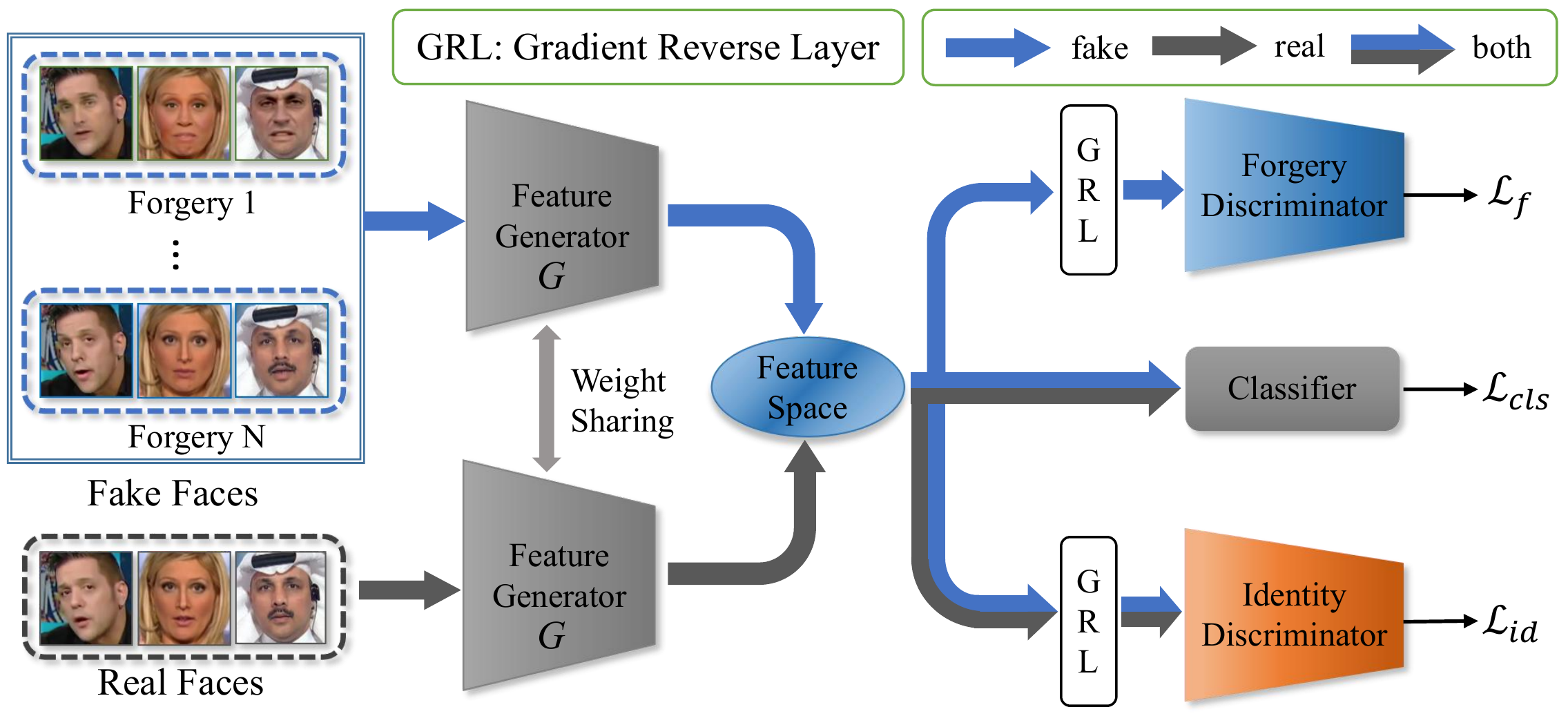}
    \caption{\small{An overview of our proposed method. With the help of adversarial learning, the feature generator extracts a feature space that is independent of forgery methods and facial identities.}}
    \label{fig:framework}
    \vspace{-12pt}
\end{figure}

Assume $\textit{X}_f$, $\textit{Y}_f$ represent forgery samples and their corresponding forgery method labels that indicate the forgery methods used for generating them. As shown in Fig.\ref{fig:framework}, the forgery method discriminator aims to distinguish the input fake faces with different forgery method labels. 
$\textit{G}$ represents the feature generator that transforms the input faces $\textit{X}$ into a latent feature embedding $\textit{Z} = \textit{G}(\textit{X})$.

We use $\textit{Z}_f$ to represent the latent features corresponding to the forgery images $\textit{X}_f$. After feature extraction, $\textit{Z}_f$ is fed into the forgery method discriminator $\textit{D}_{f}$ to determine which forgery method the input is manipulated with. 
If $\textit{D}_{f}$ is unable to distinguish samples from different forgery methods, the extracted features would capture more intrinsic common property of forgery images with various forgery types.
To achieve this goal, we introduce the forgery adversarial loss. During the training process, the parameters of the forgery method discriminator are optimized by minimizing the adversarial loss while those of feature generator are optimized by maximizing it. In our proposed method, we use the conventional adversarial loss for optimizing \textit{G} and $\textit{D}_{f}$:
\begin{equation}
\small
    \min \limits_{D_{f}} \max \limits_{G} \mathcal{L}_{f}(\theta_{G}, \theta_{D_{f}})=-\mathbbm{E}_{x,y \sim X_f,Y_f}\sum_{n=1}^{N} \mathbbm{1}_{[n=y]} \log D_{f}(G(x)),
\end{equation}
where $\theta_{G}$, $\theta_{D_{f}}$ represent the parameters of the feature generator \textit{G} and discriminator $\textit{D}_{f}$. $\mathbbm{1}_{[n=y]}$ is a one-hot vector with the y-th element set to 1. $D_{f}(G(x))$ is the N-dim vector indicating the probabilities of the extracted feature $G(x)$ being classified as each forgery method.


\vspace{-6pt}
\subsection{Identity Adversarial Learning}\label{identity section}
Since the facial identity information would also bring bias to the feature generator during the learning process, we'd like to train a feature generator that can extract discriminative features for general forgery detection while making the identities indistinguishable. To this end, we introduce the identity discriminator $\textit{D}_{id}$ which aims to distinguish the identity of the input $\textit{X}$ 
and use the adversarial learning strategy to train the generator $\textit{G}$ and the identity discriminator $\textit{D}_{id}$.

We propose two methods to provide supervision for identity discriminator training: 1) Use hard identity labels. 2)  Considering the case that we cannot always obtain ground truth hard identity labels of the dataset, we design a method to supervise $\textit{D}_{id}$ by the similarity of identity feature embeddings derived from the off-the-shelf face recognition model. Without extra explanation, the models in the experiments use the second method by default.

\vspace{-6pt}
\subsubsection{\textbf{Hard Identity Label Supervision}}
If the training samples are paired with ground truth identity labels, we can design the $\textit{D}_{id}$ as a classifier of those identity categories. The output of $\textit{D}_{id}$ is a $N_{id}$-dim vector indicating the probabilities of the input face belonging to each identity.
Feature generator \textit{G} and identity discriminator $\textit{D}_{id}$ are trained with the opposite optimization goal. We adopt conventional adversarial loss for optimizing \textit{G} and $\textit{D}_{id}$:
\begin{equation}
\small
    \min \limits_{D_{id}} \max \limits_{\textit{G}} \mathcal{L}_{id}(\theta_{G}, \theta_{D_{id}}) = -\mathbbm{E}_{x,y \sim X,Y_{id}} \sum_{n=1}^{N_{id}} {\mathbbm{1}}_{[n=y]} \log \textit{D}_{id}(G(x)),
\end{equation}
where $\textit{Y}_{id}$ represents the hard identity labels of $\textit{X}$ and $N_{id}$ is the number of facial identities. For the forgery face generated by face swapping, its identity label is set to its target face identity. In the contrast, the identity label of fake face generated by facial reenactment stays same as its source face.

\vspace{-6pt}
\subsubsection{\textbf{Identity Similarity Supervision}}
In the case that it is unable to obtain the identity label of training samples, we make use of an off-the-shelf face recognition model \textit{Arcface}\cite{deng2019arcface} to obtain the embedding features of face images. Instead of deriving identity label directly, we calculate cosine similarities between features of different face images, and adopt such similarities as supervision to train our $D_{id}$. Denote $\textit{F}_{id}$ as face recognition network. $\textit{F}_{id}(x)$ is the $512 \times 1$ dimensions embedding feature of the input sample \textit{x}. Assume $X_m$, $X_n$ representing m-th and n-th samples in a batch. We compute the identity similarity $\textit{I}_{mn}$ of two input faces $X_m$, $X_n$ using cosine distance as:
\begin{equation}
\small
    \textit{I}_{mn} = \frac{\textit{F}_{id}(X_m)^{T} \textit{F}_{id}(X_n)}{\left \| \textit{F}_{id}(X_m) \right \|_2 \left \| \textit{F}_{id}(X_n) \right \|_2},
    \textit{Y}_{mn} = 1 \mbox{ if }  \textit{I}_{mn} > \tau \mbox{ else } 0.
\end{equation}

$\textit{I}_{mn}$ is in the range of [-1,1]. $\textit{I}_{mn}$ is close to 1 when the input faces $\textit{X}_m$, $\textit{X}_n$ have similar appearance. For better utilizing $I_{mn}$ to supervise training, we further convert $I_{mn}$ into the binary label using a simple threshold $\tau$. By experimental analysis, we determine the exact value for threshold $\tau$, which could then be used to obtain the binary label $\textit{Y}_{mn}$.  Considering the distribution of identity similarity among training samples, we set $\tau=0.07$, the details of the experiment determining parameter $\tau$ are introduced in the supplementary materials.

The identity discriminator $\textit{D}_{id}$ tries to distinguish whether the difference between two extracted features can indicate their identity relationship.
It takes the element square distance between two features as input, and outputs a scaled value between 0 and 1 to represent the predicted similarity.

\begin{equation}
    \hat{\textit{Z}}_{mn}=(\textit{Z}_m-\textit{Z}_n)^2,\hat{\textit{Y}}_{mn}=\textit{D}_{id}(\hat{\textit{Z}}_{mn}),
\end{equation}
where $\textit{Z}_m$, $\textit{Z}_n$ represent the feature embedding of $\textit{X}_m$, $\textit{X}_n$ from feature generator $\textit{G}$, and $\hat{\textit{Y}}_{mn}$ is the predicted similarity between $\textit{X}_m$ and $\textit{X}_n$.

In the training process, we randomly sample the same amount of real and fake faces in a batch. Since FaceForensics++ has many identities, there are much less pairs with similar identity within a batch.
In order to balance the distribution of identity similarity, we employ focal loss as identity adversarial loss.
Assume there are M samples within a batch, we compute the adversarial loss for optimizing \textit{G} and $\textit{D}_{id}$:
\begin{equation}
    \min \limits_{D_{id}} \max \limits_{G} \mathcal{L}_{id}(\theta_{G}, \theta_{D_{id}}) = \mathbbm{E}_{x,y \sim X,Y_{id}}\sum_{\substack{m,n=1\\m<n}}^{M} \mathcal{L}_{id-mn},
\end{equation}
where $\mathcal{L}_{id-mn}$ represents the focal loss between binary label $\textit{Y}_{mn}$ and predicted similarity $\hat{\textit{Y}}_{mn}$. We formalize it as:
\begin{align}
\small
    \mathcal{L}_{id-mn}= &-[Y_{mn} \alpha (1-\hat{Y}_{mn})^\beta \log \hat{Y}_{mn}+ \notag \\ 
    &(1-Y_{mn})(1-\alpha)\hat{Y}_{mn}^\beta \log(1-\hat{Y}_{mn})].
\end{align}

We set $\alpha=0.25$ and $\beta=2$ following\cite{lin2017focal}. 
Applying identity adversarial learning forces the feature generator to generate identity-irrelevant features.

\subsection{Total Loss Function}

In order to simultaneously optimize $\theta_{G}$, $\theta_{D_{f}}$ and $\theta_{D_{\textit{id}}}$,
we apply a Gradient Reverse Layer (GRL)\cite{ganin2015unsupervised} before the forgery method discriminator and identity discriminator. During the backpropagation, the GRL takes the gradient from the subsequent layer and changes its sign before passing it to the preceding layer. In order to suppress the effect of the noisy signals at the early stage of training, we set: 
\begin{align}
\small
   \lambda = \frac{2}{1+\exp(-\gamma p))}-1,
    p=\frac{current\_iters}{total\_iters}, 
\end{align}
 in which $\gamma$ is set to 10 in all stages following\cite{ganin2015unsupervised}. What GRL actually does is to multiply the gradient of the adversarial loss by $-\lambda$ during backward propagation. 

Besides those two discriminators, the embedding features are also fed into a simple binary classifier to predict whether the input face image is real or fake. We utilize standard cross-entropy loss for the binary classifier, donated as $\mathcal{L}_{cls}$.

With above modules, the total loss is defined as:
\begin{equation}
    \mathcal{L}_{tol} = \mathcal{L}_{cls} + \lambda_{1}*\mathcal{L}_{f} + \lambda_{2}*\mathcal{L}_{id},
\label{tolloss function}
\end{equation}
where $\lambda_{1}$ and $\lambda_{2}$ are the weights for $\mathcal{L}_{f}$ and $\mathcal{L}_{id}$. We set $\lambda_{1}=0.8$ and $\lambda_2=5$. The parameter experiments determining $\lambda_{1}$ and $\lambda_{2}$ are shown in the supplementary materials.

\section{Experiments}

We use \textbf{Xception} as the feature generator following\cite{rossler2019faceforensics++}. For further implementation details and training hyper-parameter settings, please refer to supplementary materials. 

\begin{table}[!t]
\caption{\textbf{Intra-Dataset experimental results.} Quantitative video-level test results on FaceForensics++ dataset with all quality settings, i.e. LQ indicates low quality, HQ indicates high quality. Methods marked with $^{\ast}$ indicate they utilize extra frequency features.}
\small
\setlength{\tabcolsep}{0.9mm}{
\begin{tabular}{l|cc|cc}
\Xhline{1.0pt}
\multirow{2}{*}{Methods} & \multicolumn{2}{c|}{FF++.HQ} & \multicolumn{2}{c}{FF++.LQ} \\ \cline{2-5} 
 & AUC(\%) & ACC(\%) & AUC(\%) & ACC(\%) \\ \hline \hline
DSP-FWA\cite{li2018exposing} & 57.49 &——  & 62.34 &——  \\
Face X-ray\cite{li2020face}  & 87.4 &——  & 61.6 &——  \\
Xception\cite{rossler2019faceforensics++}   & 96.3 & 95.73 & 89.3 & 86.86 \\
Multi-attention\cite{Zhao_2021_CVPR} & 98.97 & 96.37 & 87.26 & 86.95 \\
F3-Net$^{\ast}$\cite{qian2020thinking}  & 98.1 & 97.52 & 93.3 & 90.43 \\
FDFL$^{\ast}$\cite{Li_2021_CVPR} & 99.3 & 96.69 & 92.4 & 89.00 \\
Inconsistency-Aware$^{\ast}$\cite{9447758} & \textbf{99.60} & 96.95 & 92.97 & 88.96 \\ \hline
\textbf{Ours} & 99.28 & \textbf{97.71} & \textbf{94.74} & \textbf{92.00} \\ \Xhline{1.0pt}
\end{tabular}
}
\label{Intra-dataset experiment}
\end{table}

\subsection{Comparsion with Other Methods}

\subsubsection{\textbf{Intra-Dataset Results}}

On the FaceForensics++\cite{rossler2019faceforensics++} dataset, we implement the identity discriminator with identity similarity supervision, and use four forgery methods (Deepfakes\cite{DeepFakes2019}, Face2Face\cite{thies2016face2face}, FaceSwap\cite{FaceSwap2019}, NeuralTextures\cite{thies2019deferred}) for the forgery method discriminator. 
We compare our method with other state-of-the-art methods at the video level. The results are reported in Table~\ref{Intra-dataset experiment} in which
most methods are frame-based detection methods.
It can be seen that the proposed method outperforms other methods in different quality settings of FaceForensics++ dataset, especially for the low-quality setting, which demonstrates the effectiveness of the proposed method regardless of the video quality. 
Note that methods marked with $^{\ast}$ utilize extra frequency features and achieve high performance on FaceForensics++. Without using extra frequency features, our method still achieves comparable results in high quality setting and outperforms them in low quality detection.

\begin{table}[t]
\caption{\textbf{Cross-dataset experimental results.} Our models are trained on FaceForensics++ and tested on Celeb-DF\cite{li2020celeb} and DeepFakeDetection\cite{DFD2019} in frame level AUC(\%). DeepFakeDetection testset consists of three compression qualities(raw, HQ, LQ).}
\small
\centering
\setlength{\tabcolsep}{0.9mm}{
\begin{tabular}{l|cc}
\Xhline{1.0pt}
Methods & DeepFakeDetection & Celeb-DF(v2) \\ \hline \hline
Xception\cite{rossler2019faceforensics++} & 68.48 & 65.5 \\
Capsule\cite{nguyen2019capsule} & 63.26 & 57.5 \\
Xception+Reg.\cite{dang2020detection} & 69.78 & 68.02  \\
Multi-Attention\cite{Zhao_2021_CVPR} & 73.18 & 67.44 \\
Inconsistency-Aware$^{\ast}$\cite{9447758} &——  & 72.3 \\
Two-branch$^{\ast}$\cite{masi2020two} &——  & \textbf{73.41} \\ \hline
\textbf{Ours} & \textbf{79.69} & 72.77 \\ \Xhline{1.0pt}
\end{tabular}
}
\label{Cross-dataset experimental results}
\vspace{-8pt}
\end{table}



\begin{figure}[!t]
\centering
\makebox[\linewidth][c]{
\subfigure[adv-forgery]{
\includegraphics[width=0.47\linewidth]{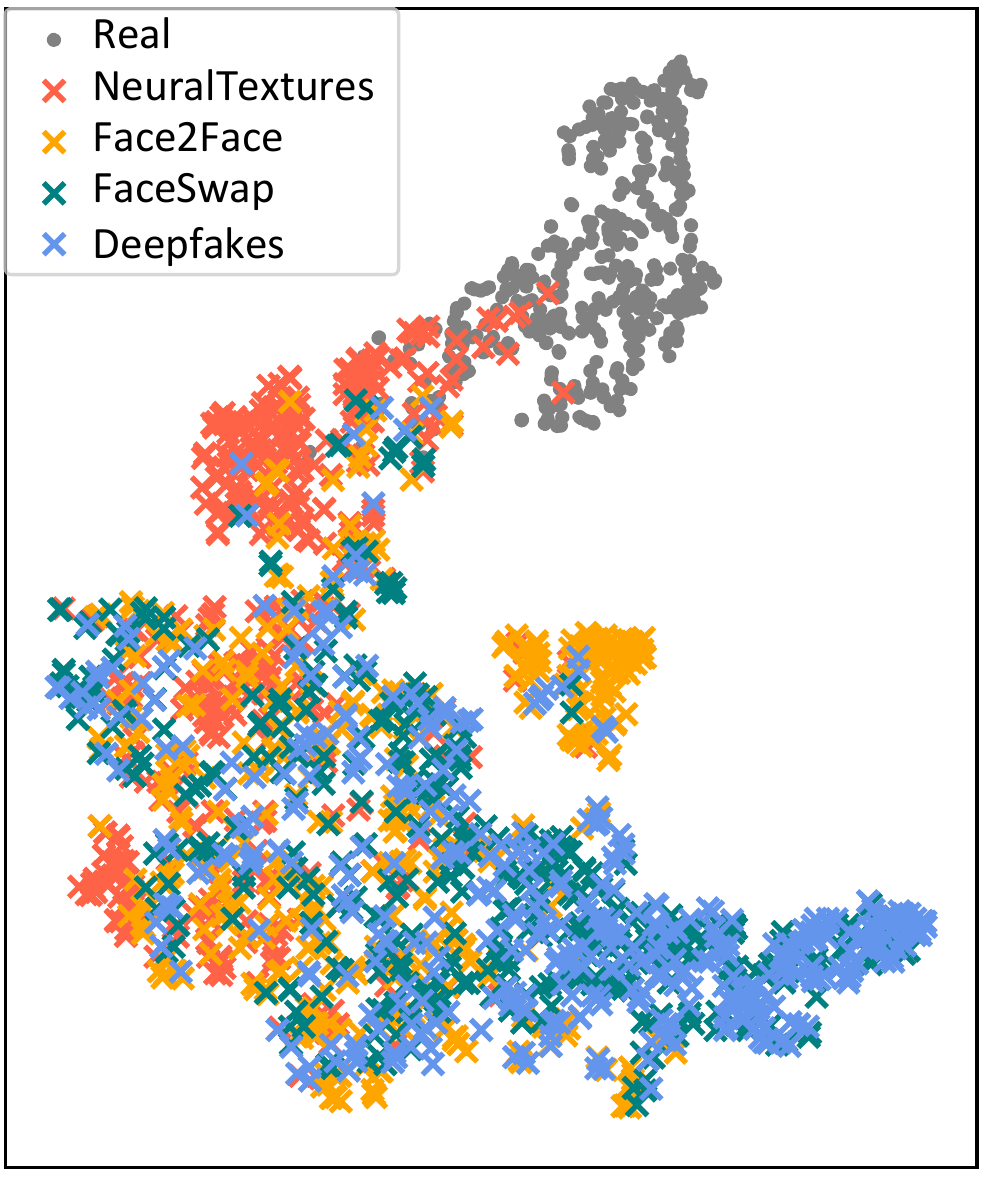}
\label{ad-fake}
}\hspace*{-0.6em}
\subfigure[adv-identity]{
\includegraphics[width=0.47\linewidth]{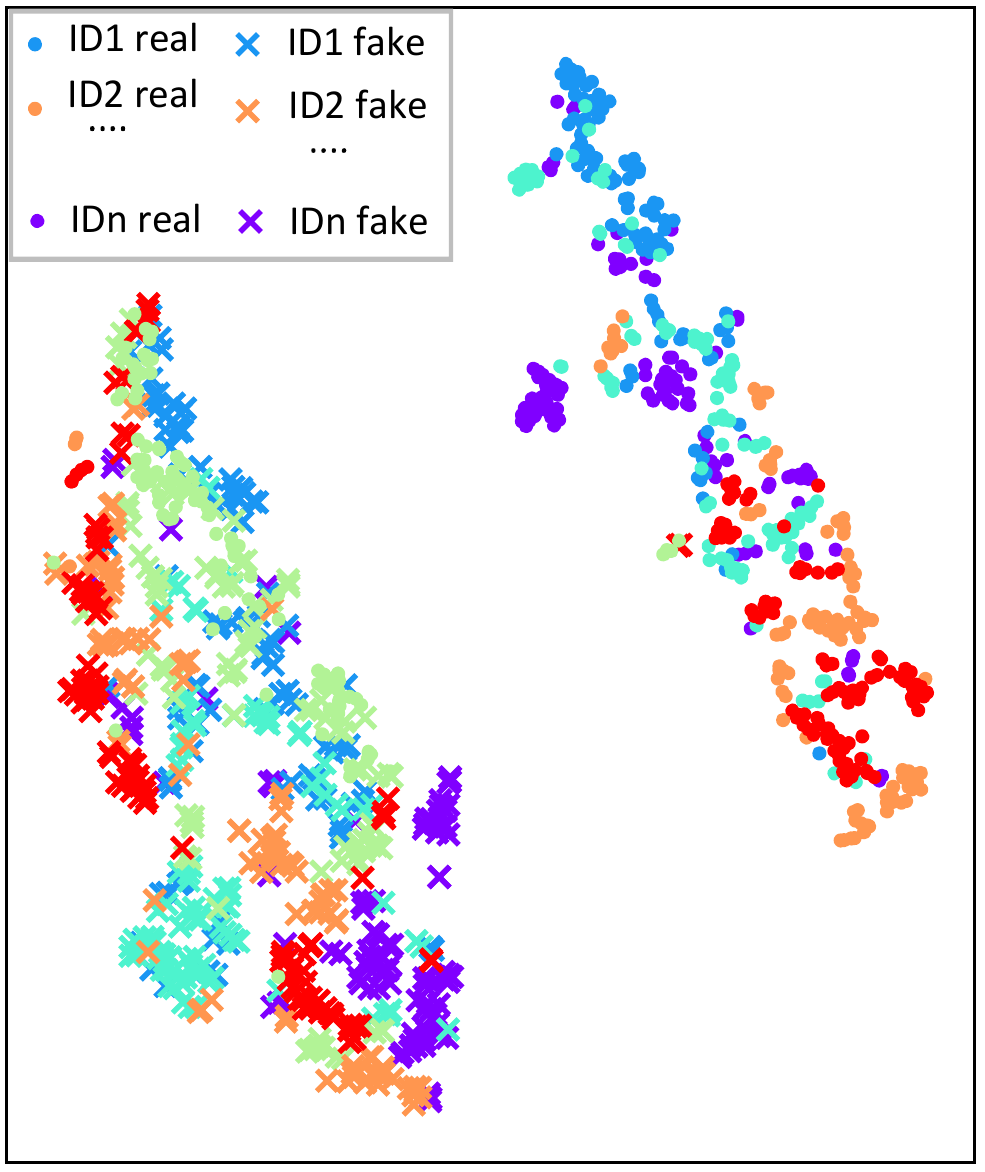}
\label{ad-id}
}}
\caption{The t-SNE visualization on FaceForensics++ low quality task. (a) Baseline with adversarial forgery method discriminator. 
(b) Baseline with adversarial identity discriminator. }
\label{Intra-dataset figure}
\vspace{-12pt}
\end{figure}

%
\vspace{-6pt}
\subsubsection{\textbf{Cross-Dataset Results.}}
For verifying the generalization ability across datasets of the proposed method, we evaluate performances on Celeb-DF while training on FaceForensics++. 
As shown in Table~\ref{Cross-dataset experimental results}, our method achieves superior performance than other methods except for Two-branch. Note that Two-branch is a video-based method that utilizes temporal information. Although without temporal information, our method is just slightly inferior than Two-branch. These experimental results demonstrate the strong generalization ability of proposed method.

We also evaluate the generalization ability over DeepFakeDetection dataset, which contains thousands of facial fake videos of three quality levels (raw, high quality, low quality). Because there is no official standard testing set, we randomly partition a testing set containing 150 manipulated videos and 150 original videos. Each video of three compression qualities is tested at the frame level. The result reports a clear improvement over the other methods, and illustrates that our proposed method could learn intrinsic common discriminative features and thus improves the generalization ability.

\subsection{Ablation Study}
\subsubsection{The Effectiveness of the Proposed Components}

We illustrate the t-SNE visualization of the features extracted from the model equipped with the forgery method adversarial learning or the identity adversarial learning, denoted as adv-forgery and adv-identity respectively in Fig.~\ref{Intra-dataset figure}.
As shown in Fig.~\ref{ad-fake} and Fig.\ref{ad-id}, the distribution of the fake faces features with different forgery methods and the distribution of the features with different identities are both dispersed. Besides, the feature distribution of the real faces or the fake faces is more compact. It verifies the effectiveness of the proposed adversarial learning components on eliminating the effect of the face forgery methods and the facial identities. We also attempt to utilize some classic clustering methods to cluster features extracted by the Feature Generator and use the clustering accuracy as the metric. Detail experimental results are introduced in the supplementary materials.

\begin{table}[!t]
\caption{\textbf{Ablation Study.}  We make the evaluations of different components of the proposed method in intra-dataset (FaceForensics++) and cross-dataset (Celeb-DF(v2)) settings.}
\vspace{-15pt}
\begin{center}
\small
\renewcommand{\arraystretch}{1.0}
\begin{tabular}{c|cc|cc}
\Xhline{1.0pt}
    \multirow{2}{*}{Model}            & \multicolumn{2}{c|}{FaceForensics++} & \multicolumn{2}{c}{Celeb-DF(v2)}     \\ \cline{2-5} 
              & AUC(\%)             & ACC(\%)            & AUC(\%)        & ACC(\%)         \\ \hline \hline
baseline     & 91.93      & 89.29               & 65.76            & 67.45                    \\ 
adv-forgery     & 93.10     & 90.14               & 68.96         & 68.90                 \\ 
adv-identity  & 93.63    & 91.00               & 71.02              & 68.48                 \\ 
adv-both   & \textbf{94.74}   & \textbf{92.00}    & \textbf{72.77}   & \textbf{70.34}\\ \Xhline{1.0pt} 
\end{tabular}
\end{center}
\label{Ablation Study}
\vspace{-15pt}
\end{table}

\begin{table}[!t]
\centering
\small
\caption{\textbf{Experimental results on unseen forgery method.} Video-level ACC and AUC when testing on each forgery method of FaceForensics++.LQ after training on the remaining four.}
\renewcommand{\arraystretch}{1.0}
\setlength{\tabcolsep}{0.5mm}{
\begin{tabular}{c|c|c|cc}
\Xhline{1.0pt}
Train set & Test set                        & Methods  & ACC(\%)         & AUC(\%)         \\ \hline \hline
\multirow{10}{*}{\begin{tabular}[c]{@{}c@{}}Trained on \\the remaining\\four forgery\\dataset\end{tabular}} & \multirow{2}{*}{Deepfakes} & Xception & 83.214 & 89.872 \\
          &                                 & Ours     & \textbf{85.00}     & \textbf{90.96} \\ \cline{2-5} 
          & \multirow{2}{*}{Face2Face}      & Xception & 59.64          & 69.62          \\
          &                                 & Ours     & \textbf{62.50}   & \textbf{73.92} \\ \cline{2-5} 
          & \multirow{2}{*}{FaceSwap}       & Xception & 59.29          & 70.38          \\
          &                                 & Ours     & \textbf{63.21} & \textbf{74.04} \\ \cline{2-5} 
          & \multirow{2}{*}{NeuralTextures} & Xception & 58.93          & 65.75          \\
          &                                 & Ours     & \textbf{61.07} & \textbf{70.04} \\ \cline{2-5} 
          & \multirow{2}{*}{FaceShifter}    & Xception & 71.79          & 79.15          \\
          &                                 & Ours     & \textbf{74.64} & \textbf{80.31} \\ \Xhline{1.0pt}
\end{tabular}
}
\label{Cross-manipulation}
\vspace{-15pt}
\end{table}

In Table.\ref{Ablation Study}, we show the face forgery detection performance of models with different proposed components. All the models are trained on FF++ low quality training set and evaluated on both its testing set and Celeb-DF(v2).
The experimental results show that both adversarial learning components can not only improve the performance of face forgery detection in the intra-dataset setting but also achieve better generalization ability across dataset compared to the baseline, which demonstrates the effectiveness of the two components. What's more, the model equipped with both components, denoted as adv-both in Table.~\ref{Ablation Study}, can obtain further improvements, which demonstrates that the two adversarial learning components are complementary to each other.

\vspace{-6pt}
\subsubsection{Generalize to Unseen Forgery Method}
To assess the generalization ability to unseen forgery method, we utilize the newly released FaceForensics++ dataset that has fake videos created by five different forgery methods: DeepFakes, Face2Face, FaceSwap, NeuralTextures and FaceShifter. FaceShifter\cite{Li_2020_CVPR} is a recently proposed two-stage face swapping method which is newly added to FaceForensics++.
In Table.\ref{Cross-manipulation}, we test on each forgery method data using the model trained on the remaining four forgery methods in the low quality setting. The results show that our method consistently achieves superior performance compared to baseline, which demonstrates the better generalization ability of the proposed method to unseen forgeries.

\begin{figure}[!t]
    \centering
    \includegraphics[width=0.8\linewidth]{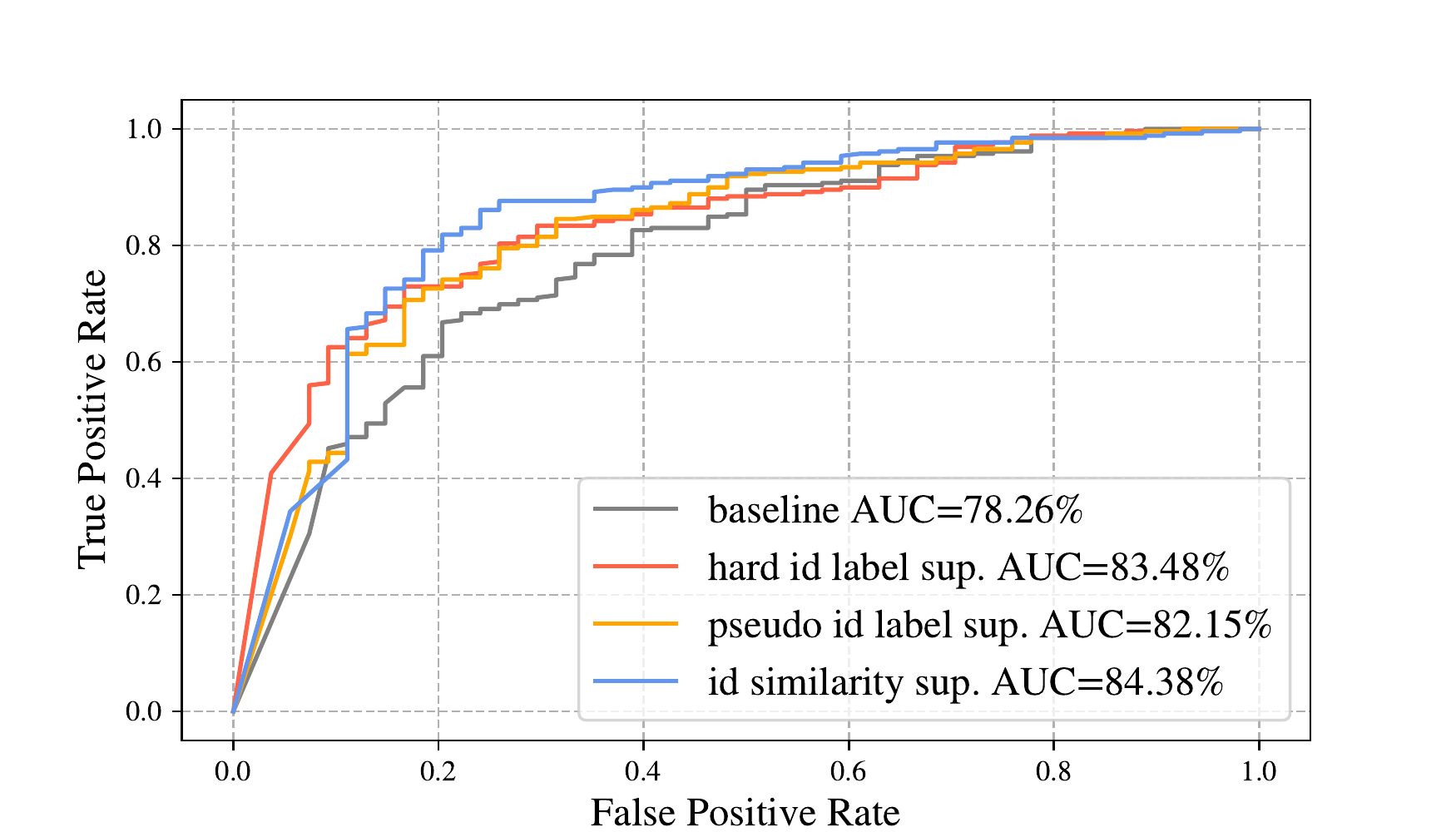}
    \caption{The ROC performance of different identity supervision.}
    \label{ROC figure}
    \vspace{-5pt}
\end{figure}

\vspace{-6pt}
\subsubsection{The Performance of Different Identity Supervision}
In Section \ref{identity section}, we introduce two approaches to implement identity discriminator. 
Using the pseudo identity labels according to the clustering results for training is another possible approach.
Here, we conduct experiments and compare the performance of these different approaches on the DFD dataset, since it has the ground truth label of facial identities. 
For pseudo identity label supervision, we adopt K-Means to produce the pseudo identity labels with the number of the clusters set to the same as the ground truth number of identities.
As shown in Fig.~\ref{ROC figure}, all three approaches show superior performance compared to baseline, which again demonstrates the effectiveness of the identity adversarial learning. As expected, the pseudo identity label supervision performs worse than the hard identity supervision, due to the inaccurate clustering results.
Surprisingly, the identity similarity supervision outperforms the hard identity label supervision. We blame it on the fact that the hard identity labels of fake faces produced by face swapping methods are assigned to be the same as their corresponding target faces, which may be incorrect due to imperfect performance of face swapping methods.

\vspace{-6pt}
\section{Conclusion}
In this work, we propose a novel adversarial learning method for face forgery detection, which aims to eliminate the effect of forgery methods and facial identities for learning intrinsic common discriminative features. Through two adversarial discriminators for forgery methods and facial identities respectively, our network generates a compact feature space that is independent of forgery methods and facial identities.
In the case of lacking ground truth label of facial identities, we propose a special identity discriminator based on similarity information derived from off-the-shelf face recognition model. Extensive experiments demonstrate the effectiveness of the proposed method in intra-dataset and cross-dataset settings.
\vspace{-6pt}
\section{ACKNOWLEDGEMENT}
This work is supported by the National Natural Science Foundation of China (No. 62002336, No. U20B2047) and Exploration Fund Project of University of Science and Technology of China under Grant YD3480002001.

\appendix
\section{Implementation Details}
\subsection{Network Structure}
We use \textbf{Xception} as the feature generator following \cite{rossler2019faceforensics++}, which maps the input images into 2048-D embedding feature vectors in latent representation space. The binary classifier consists of a fully connected layer with two output nodes representing real vs. fake. Given samples manipulated with \textit{N} forgery methods as input data, the forgery method discriminator consists of three fully connected layers with 512, 512, and \textit{N} units respectively. The identity discriminator also consists of three fully connected layers with 512, 512, and $\textit{N}_{id}$ units for the case of identity label supervision, while with 512, 512 hidden units and 1 output node when supervised with identity similarity.



\subsection{Training Details}
Following \cite{thies2016face2face}, we employ the open source dlib algorithm to do face detection and landmark localization. All the detected faces are cropped (enlarged by a factor of 1.3) around the center of the face, and then resized to $299 \times 299$. 
We initialize the feature generator with parameters pre-trained on ImageNet \cite{deng2009imagenet}. In the training process, the batch size is set to 64 and the maximum total number of iterations is set to 20,000.  
We adopt Adam optimizer and the initial learning rate is set to 0.0001 and then gradually descends to nearly 0. 
During training, we compute validation AUC and ACC ten times per epoch and stop the training process if the validation accuracy does not change for 20 consecutive checks. 
Following FaceForensics++ \cite{rossler2019faceforensics++}, we use 110 frames per videos for testing. 
\section{Supplementary Experiments}
\subsection{The Effectiveness of the Proposed Components}

To further verify the effectiveness of adversarial learning components, we attempt to utilize some classic clustering methods to cluster features extracted by the Feature Generator and use the clustering accuracy as the metric. The lower clustering accuracy indicates less correlation.
Table.~\ref{cluster forgery} and Table.~\ref{cluster identity} shows the clustering accuracy of baseline method and our method using K-Means and GaussianMixture for forgery method clustering and identity clustering. 
For forgery method clustering in Table.~\ref{cluster forgery}, the features are extracted from FaceForensics++. Compared to the baseline, the clustering accuracy of our method drops off 19.47\% (using K-Means) and 21.07\% (using GaussianMixture) respectively. For identity clustering in Table.~\ref{cluster identity}, the features are extracted from the real faces of DeepFakeDetection(DFD) dataset which has identity annotations. Compared to the baseline, the clustering accuracy of our method drops off 7.72\% (using K-Means) and 8.54\% (using GaussianMixture) respectively. These clustering results further demonstrate that the features learned by the proposed method are less related to different forgery methods and facial identities. 

\begin{table}[!t]
\caption{\textbf{Forgery method clustering accuracy.} The forgery method clustering accuracy of features from four forgery methods (Deepfakes, Face2Face, FaceSwap, NeuralTextures) of FaceForensics++.}
\small
\centering
\renewcommand{\arraystretch}{1.0}
\setlength{\tabcolsep}{5mm}{
\begin{tabular}{c|cc}
\Xhline{1.0pt} 
\multirow{2}{*}{ACC(\%)}  & \multicolumn{2}{c}{Forgery Method Clustering Accuracy} \\ \cline{2-3}
                         & K-Means         & GaussianMixture        \\ \hline \hline
Xception                 & 49.29           & 51.25                  \\ \hline
Ours                     & 29.82           & 30.18                  \\ \Xhline{1.0pt}
\end{tabular}}
\label{cluster forgery}
\end{table}

\begin{table}[!t]
\centering
\small
\caption{\textbf{Identity clustering accuracy.} The identity clustering accuracy of features from DeepFakeDetection(DFD) original videos with 28 facial identities.}
\renewcommand{\arraystretch}{1.0}
\setlength{\tabcolsep}{5.5mm}{
\begin{tabular}{c|cc}
\Xhline{1.0pt}
\multirow{2}{*}{ACC(\%)}  & \multicolumn{2}{c}{Identity Clustering Accuracy} \\ \cline{2-3}
                        & K-Means & GaussianMixture \\ \hline \hline
Xception                & 30.03   & 32.51           \\ \hline
Ours                    & 22.31   & 23.97           \\ \Xhline{1.0pt}
\end{tabular}}
\label{cluster identity}
\end{table}


\subsection{Hyper-Parameter Experiments}

\subsubsection{Adversarial loss weights $\lambda_1$, $\lambda_2$}
\begin{figure}[h]
    \centering
    \includegraphics[width=\linewidth]{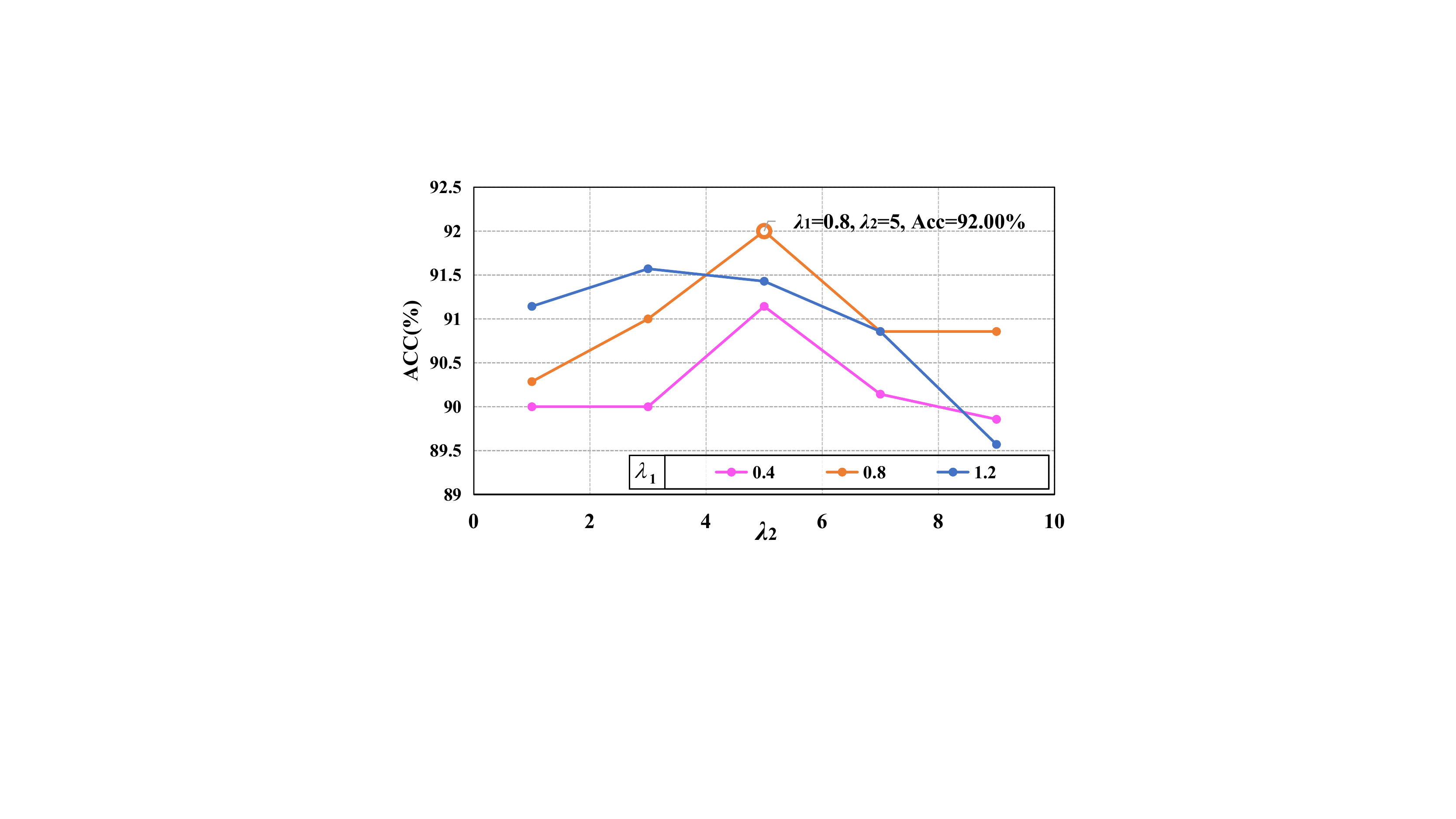}
    \caption{The detection ACC(\%) with different loss weights $\lambda_1$, $\lambda_2$.}
    \label{paremeter_lambda}
\end{figure}
The total loss is defined as:
\begin{equation}
    \mathcal{L}_{tol} = \mathcal{L}_{cls} + \lambda_{1}*\mathcal{L}_{f} + \lambda_{2}*\mathcal{L}_{id}.
\label{tolloss function}
\end{equation}

We explore the appropriate weights for adversarial loss in Eq.~\ref{tolloss function}. As shown in Fig.~\ref{paremeter_lambda}, we set a series of adversarial loss weights $\lambda_1$, $\lambda_2$, and compare their performance on the FaceForensics++ low-quality detection task. We can see that the performance curve shows a trend of rising first and then falling with the fixed $\lambda_1$ and the best detection accuracy is achieved when $\lambda_1$ and $\lambda_2$ are set as 0.8 and 5.

\subsubsection{Similarity Threshold $\tau$}

\begin{figure}[!t]
    \centering
    \includegraphics[width=\linewidth]{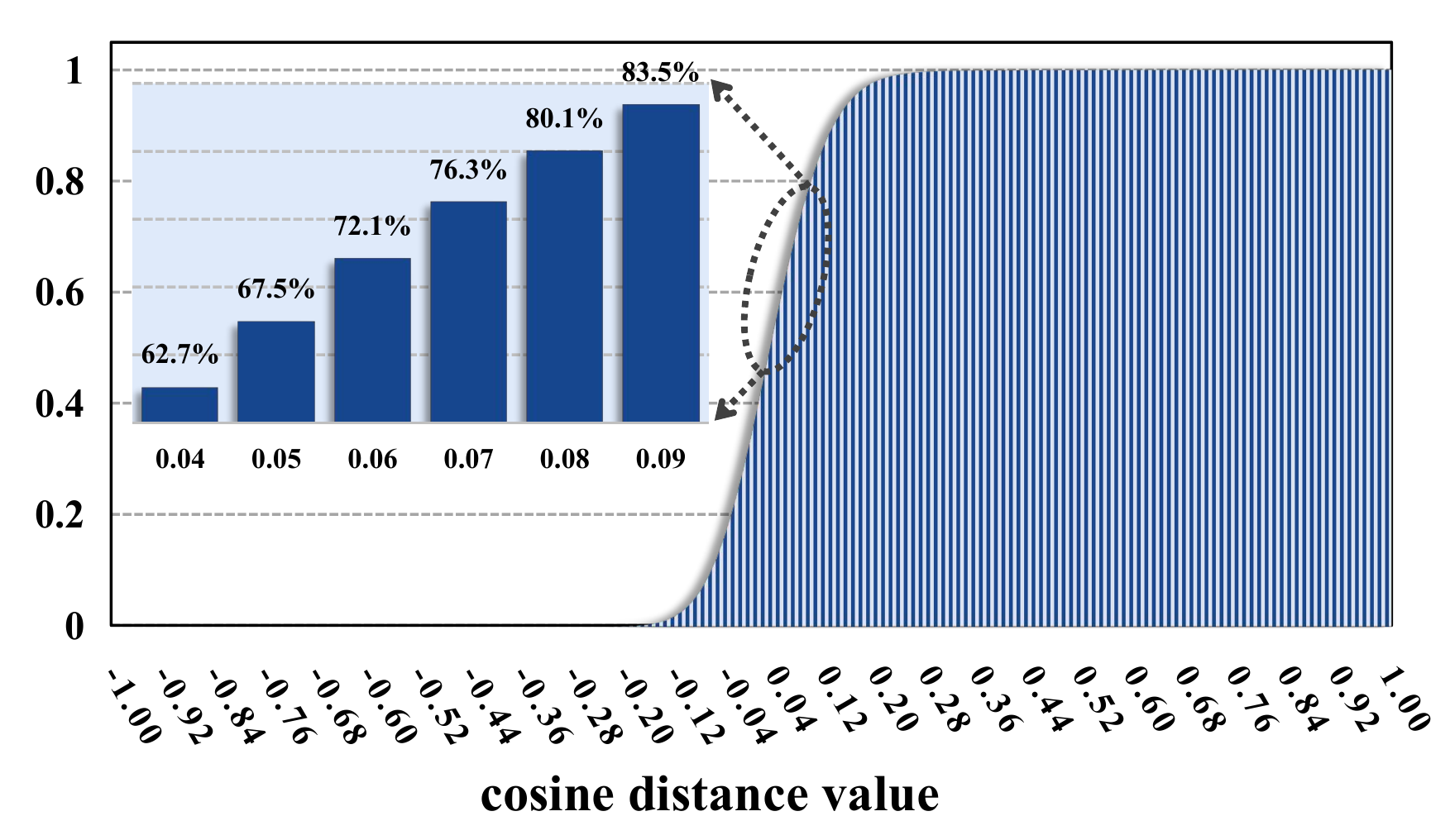}
    \caption{The cumulative distribution function of the sampled cosine distances. The bar figure describes the fine statistic distribution in the range [0.04, 0.09].}
    \label{distribution}
\end{figure}

We first limit the range of threshold $\tau$ through observing the distribution of facial similarity of the training samples. Specifically, we randomly sample 100 batches and compute the cosine distance between any two samples within a batch. Then we calculate the cumulative distribution of these cosine distances to approximate the distribution of facial identity similarity of train data. Fig.~\ref{distribution} shows the cumulative distribution curve. We roughly choose the cumulative probability interval from 60\% to 85\%, since there are less samples with similar identity than dissimilar. According to the cumulative distribution curve, we limit the range of the threshold $\tau$ to [0.04, 0.09] and then experimentally compare the performance with different values of $\tau$ in the range of [0.04, 0.09], where the step size is set to 0.01. 
\begin{figure}[t]
    \centering
    \includegraphics[width=\linewidth]{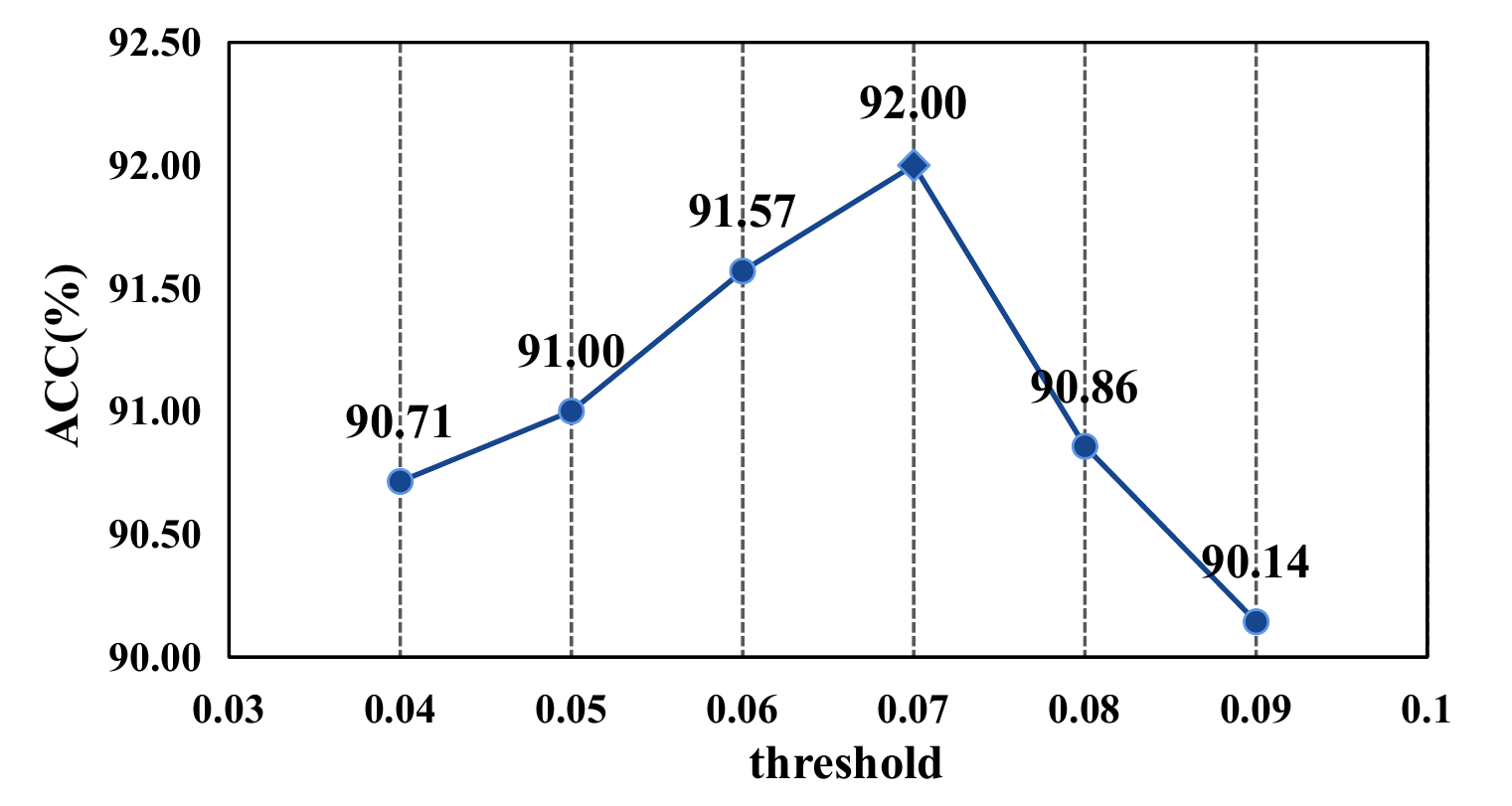}
    \caption{The ACC performance in video level of a series of $\tau$.}
    \label{threshold}
\end{figure}

The experimental results are shown in Fig.~\ref{threshold}. It can be seen that the best performance on FaceForensics++ low quality task when $\tau$ is set to 0.07.

\end{document}